%% file: neurips_2026.tex
\documentclass{article}

 \usepackage[preprint]{neurips_2026}

\usepackage[utf8]{inputenc} 
\usepackage[T1]{fontenc}    
\usepackage{hyperref}       
\usepackage{url}            
\usepackage{booktabs}       
\usepackage{amsfonts}       
\usepackage{nicefrac}       
\usepackage{microtype}      
\usepackage{xcolor}         
\usepackage{amsmath}
\usepackage{amssymb}
\usepackage{mathtools}
\usepackage{amsthm}
\usepackage{algorithm}
\usepackage{algorithmic}
\usepackage{graphicx}
\usepackage{subcaption}
\usepackage{xspace}

\newcommand{\bench}[0]{Disempower-Grid\xspace}

\theoremstyle{plain}
\newtheorem{theorem}{Theorem}[section]
\newtheorem{proposition}[theorem]{Proposition}

\newtheorem{corollary}[theorem]{Corollary}
\theoremstyle{definition}
\newtheorem{definition}[theorem]{Definition}

\theoremstyle{remark}

\title{When Assisting One Disempowers Another}

\author{%
  Claire Yang\\
  University of Washington\\
  \texttt{claireyy@uw.edu} \\
  \And
  Claire Jie Zhang \\
  University of Washington\\
  \texttt{claizhan@uw.edu} \\
  \AND
  Maya Cakmak \\
  University of Washington\\
  \texttt{mcakmak@uw.edu} \\
  \And
  Max Kleiman-Weiner \\
  University of Washington\\
  \texttt{maxkw@uw.edu} \\
}

\begin{document}

\maketitle

\begin{abstract}
Personal AI agents are increasingly deployed in shared environments, where their actions affect not just the primary user they are assisting, but bystanders who never consented to being affected by the system. We show that a well-meaning AI assistant optimizing for one user's benefit can unintentionally erode a bystander's agency, a phenomenon we formalize as bystander disempowerment. We theoretically characterize the conditions under which disempowerment arises, showing it emerges when an assistant systematically selects actions that increase user empowerment at the bystander's expense. We empirically demonstrate this in Disempower-Grid, a parameterized suite of multi-agent gridworld environments, finding that between 27–96\% of procedurally generated environments exhibit disempowerment, and that the presence of disempowerment depends strongly on assistant objective and capability, not just environmental structure.

\end{abstract}

\section{Introduction}

Personal AI agents and robots are increasingly deployed in shared environments, where their actions affect not just the primary user they are assisting, but others present in the same space. A nurse's robotic assistant may inadvertently obstruct a patient; a developer's coding assistant may suggest infrastructure changes that unintentionally break other developers' workflows; a teacher's classroom AI may fail to accommodate students with learning disabilities, causing them to fall behind. Despite this, alignment research has largely focused on the case in which a single agent serves a single user in an isolated setting. This framing is insufficient. Aligning AI in a multi-agent setting is a dynamic process, where individual and collective well-being must be balanced \citep{carichon2025coming}. 

In the real world, space and resources are finite, causing physical or informational bottlenecks to naturally emerge. In this setting, AI assistants acting on behalf of a user may influence the world in such a way that others lose their ability to reach their desired states. 
 For example, an AI scheduling assistant for a manager might optimize the manager's calendar by repeatedly reserving scarce meeting rooms or preferred time slots, making it harder for other team members to coordinate their own work. The assistant need not know or intend to interfere with those bystanders; it may simply treat their lost flexibility as part of the environment.
One challenge in mitigating these negative side-effects is that the objectives of the bystanders are often unknown to the assistant and can vary widely. 

\begin{figure}[t!]
    \centering
    \includegraphics[width=1.0\linewidth, trim={0cm 0.25cm 0 0.26cm},clip]{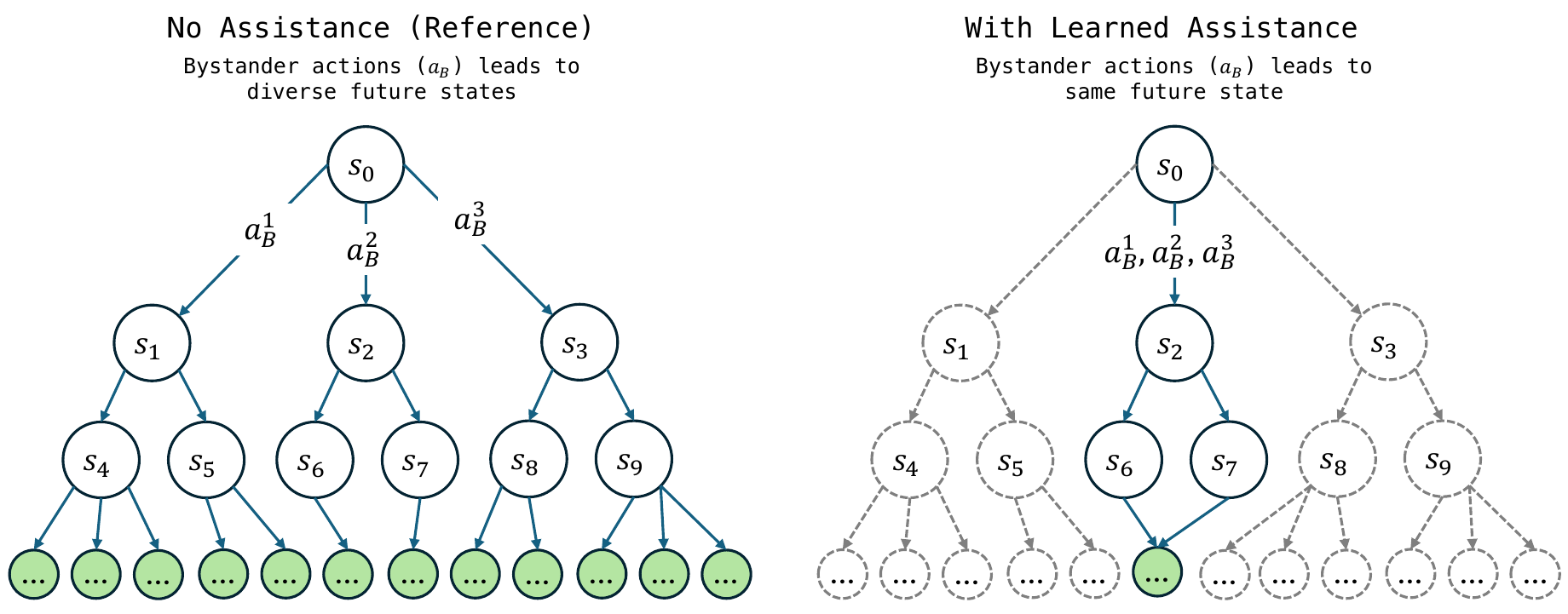}
    \caption{An example of disempowerment: Bystander actions ($a_B$) lead to diverse future states under no assistance (reference) policy (left), but converge to the same future state under the learned assistant policy (right). Disempowerment captures the bystander's loss of influence under the learned assistant policy, compared to the reference policy. It does so by measuring the difference in mutual information between bystander actions and future states. Dashed nodes indicate states no longer reachable through bystander actions under the learned policy; solid nodes indicate states the bystander can still reach through its actions. We show that assistance to a user can cause bystander disempowerment.}
    \label{fig:figure_1}
\end{figure}

To make this loss of bystander influence precise, we build on the notion of empowerment, a measure of an agent’s ability to control its environment \citep{klyubin_empowerment_2005}. We calculate changes in the bystander’s empowerment as a counterfactual measure of how the assistant’s policy affects the bystander’s agency. We propose that an assistant disempowers a bystander when its learned policy decreases the bystander’s empowerment relative to a reference assistant policy. Figure~\ref{fig:figure_1} illustrates the key distinction: under an unbiased reference assistant policy (e.g., a random uniform assistant policy), different bystander actions lead to different future states; under the learned assistant policy, those same bystander actions collapse onto the same future state. The bystander may still occupy the environment, but its actions have less influence over what happens next.

Importantly, disempowerment is task-agnostic and captures both short-term and long-term effects caused by the assistant. An assistant that influences the world repeatedly in small ways to assist the user and inadvertently decrease the bystander's agency not only affects the bystander's short-term goal achievement, but also their ability to access future opportunities.
This makes disempowerment complementary to existing measures of harm and side effects: it captures losses in a bystander’s action-conditioned influence that may be invisible to metrics based only on outcomes or state reachability \citep{krakovna_penalizing_2019, turner_conservative_2020,counterfactual_harm}

Critically, the alignment problem of bystander disempowerment need not emerge from any malicious intent. An AI agent that inadvertently disempowers others could lead to ``gradual disempowerment,'' where human agency erodes over time \citep{hammond2025multi, kulveit2025gradual}. At a societal level, it might mean one group being empowered at the expense of another, concentrating power in the hands of only a few \cite{chan2023harms}. Ensuring alignment in multi-agent settings requires explicit attention to disempowerment, even when there is only one primary user. Our contributions are as follows:

\begin{enumerate}
\item We formalize bystander disempowerment as a counterfactual decrease in a bystander's empowerment under a learned assistant's policy and theoretically characterize the conditions under which it arises, showing it emerges when an assistant systematically selects actions that benefit the user at the bystander's expense.
\item We introduce Disempower-Grid, a parameterized suite of multi-agent gridworld environments spanning diverse assistant embodiments, action spaces, and environment dynamics, designed to surface and benchmark bystander disempowerment.
\item Using Disempower-Grid, we empirically demonstrate that bystander disempowerment is a general phenomenon, occurring in 27–96\% of procedurally generated environments across conditions.
\item We show that disempowerment is policy-dependent rather than a static property of the environment. The rate of disempowerment varies dramatically across assistant objectives even when layouts are held fixed, and greater assistant capability does not monotonically increase disempowerment, suggesting that the right objective can allow a more capable assistant to help the user with less unintentional harm to the bystander.
\end{enumerate}

\section{Related Work}

We combine key ideas from goal-agnostic objectives and connect them to assistance and AI safety. 

\paragraph{Goal-Agnostic Assistance: Empowerment, Choice, and Power.}
Our work builds on key ideas from reinforcement learning and control that aim to measure an agent's control and capability in an environment. \textit{Empowerment}, defined as the maximum mutual information between an agent's action and its future states, is a goal-agnostic measure of capability \citep{klyubin_empowerment_2005, klyubin_all_2005}. An agent's effective empowerment (the mutual information, not the maximum of the mutual information) has been used as an intrinsic motivation for reinforcement learning agents, and shown to enhance their learning and exploration across domains \citep{brandle_empowerment_2023, baddam_search_2025, lidayan_intrinsically-motivated_2025}. It has also been applied to improve agent coordination in multi-agent settings \citep{van2020robust, woojun_variation_multiagent, guckelsberger2016intrinsically}. Intuitively, effective empowerment measures an agent's potential to navigate efficiently through a state space. Agents with greater mastery and control over their environment or those that can access a larger fraction of available states will have higher effective empowerment. For example, if two agents are locked in two separate rooms, the agent with a key to get out would have higher effective empowerment than the one without, since the agent with a key would also be able to potentially access states beyond the locked room (even if it does not choose to do so). Finally, Turner and Tadepalli \cite{turner2022parametrically} demonstrate that reinforcement learning-based agents are power-seeking (as measured by increases in optionality), suggesting that the majority of reward functions reward maximizing future choices \citep{turner_optimal_2023}. In this work, when we refer to empowerment henceforth, we refer to effective empowerment.

Recent work uses approximations of effective empowerment as an objective for assistance. Importantly, these models can help human users without needing to model their goals \citep{du_ave_2020, myers_learning_2024}. The appeal is intuitive: by maximizing a human's effective empowerment, an agent should help them achieve as many possible states in the future without needing to explicitly infer those goals. Franzmeyer et al. \cite{franzmeyer_learning_2022} develop an assistive agent that optimizes the number of \textit{choices} available to another agent, a simpler computation that only depends on the agent's states, not their actions. Because calculating empowerment is computationally intractable in high-dimensional environments, several approximations have been developed to scale its measurement \citep{mohamed_variational, myers_learning_2024, jung_empowerment_2012}. 

Regardless, prior works on goal-agnostic assistance focus on dyadic interactions between an assistant and a simulated human user, or assume that the user and bystanders are adversaries \citep{du_ave_2020, myers_learning_2024, franzmeyer_learning_2022}. Notably, multi-principal assistance games is a line of work that studies assistance in multi-agent environments, but they assume that the assistant aims to serve multiple users with diverging objectives. This differs from our work, in which the assistant only serves a single user in the presence of multiple bystanders with diverse objectives. Aside from being used as a measurement for agent capability and a training objective for reinforcement learning, empowerment has also been used to measure the power of large language-model agents \cite{song2025estimating}, inspiring our approach to use negative changes in empowerment as a measure for reduced human agency. 

\paragraph{Side Effects and Harm.}

There is a rich literature on studying the unintended side effects of AI action and assistance, typically focusing on passive aspects of the environment (e.g., broken or unreachable objects) \citep{amodei_concrete_2016, krakovna_penalizing_2019, turner_conservative_2020, krakovna_avoiding_2020}. This differs from our focus on side effects experienced by a bystander agent active in the environment. Harm has been proposed as a value-laden measure for negative side effects on humans, measured by comparing how well off the human is under an AI agent acting versus not acting \cite{counterfactual_harm, characterizing_manipulation}. This counterfactual approach has also been proposed for penalizing side effects through a stepwise relative reachability measure \cite{krakovna_penalizing_2019}. We take inspiration from this line of work in defining our disempowerment metric as a counterfactual measure. However, disempowerment differs from existing side-effect measures in two important respects.

 Measures such as relative reachability \citep{krakovna_penalizing_2019} and Attainable Utility Preservation (AUP) \citep{turner_conservative_2020} abstract away from agents' policies, focusing instead on the accessibility structure of the state space or the agent's own ability to satisfy auxiliary objectives. Empowerment, by contrast, measures the mutual information between an agent's actions and its resulting future states over a horizon \citep{klyubin_empowerment_2005, salgeEmpowermentIntroduction2013}, explicitly conditioning on the bystander's action choices. This distinction matters, as accessibility in the state space can still exist while being entirely facilitated by other agents or external forces. An assistant that systematically reduces the influence of the bystander's actions without fully eliminating access to those states would register no change under reachability-based measures, yet would produce measurable disempowerment. Figure~\ref{fig:figure_1} illustrates this. Under the learned assistant policy (b), dashed nodes indicate states that remain in the state space but are no longer reachable through the bystander's own actions. This makes disempowerment a more expressive metric than existing snapshot measures of harm.

\section{Preliminaries}\label{sec:prelim}
We consider a multi-agent assistance setting \cite{hadfield-menell_cooperative_2016}, with an assistive AI agent (\textbf{A}) aiming to assist a (simulated) human user (\textbf{U}) with an additional (simulated) human bystander (\textbf{B}), an agent who is not the target of assistance but can take actions in the environment. This models realistic scenarios where personal AI assistants operate in shared environments. This setting is represented as a multi-agent MDP $M = (S, \Omega_U, \Omega_B, \Omega_A, A_U, A_B, A_A, P, R_U, R_B, R_A, \gamma)$, where the states are represented by $S$, consisting of the joint states of the user, bystander, and assistant, goals, and any task-relevant environment features. The assistant's observation function $\Omega_A$ does not include the goals of the user or bystander, as the usage of empowerment enables the assistant to avoid needing to know or infer the humans' underlying intentions. The user and bystander follow goal-directed policies $\pi_U$ and $\pi_B$.
The assistant policy $\pi_A$ is learned. We treat $\pi_U$ and $\pi_B$ as fixed to isolate the effect of the assistant's learning on bystander empowerment during evaluation. They may be goal-directed policies or A* planning policies, but we do not model the bystander as learning a strategic best response to the assistant over repeated interactions. While real humans may adapt their behavior in response to repeated disempowerment, this simplifying assumption allows us to establish that disempowerment can arise even in the most straightforward multi-agent setting, providing a foundation for future work on disempowerment incorporating adaptive human models.

\paragraph{Effective Empowerment.}

For an agent $i\in\{U,B\}$, we define the $H$-step effective empowerment $E$ at state $s$ under assistant policy $\pi_A$ as

\[
E_i^H(s;\pi_A)
=
I(A^i_{0:H-1};S_H\mid S_0=s),
\]

where the mutual information is computed under the trajectory distribution induced by $(\pi_U,\pi_B,\pi_A)$. This is the effective empowerment quantity used in prior work \citep{du_ave_2020, myers_learning_2024}. 

\paragraph{Rollout-Averaged Empowerment.}

Because the assistant policy affects the states visited over an episode, we evaluate empowerment along the induced trajectory. For $i\in\{U,B\}$, define

\[
J_i^E(\pi_A)
=
\mathbb{E}_{\pi_A,\pi_U,\pi_B}
\left[
\sum_{t=0}^{\infty}
\gamma^t E_i^H(S_t;\pi_A)
\right].
\]

This infinite-horizon expression is the standard discounted objective; in our finite-horizon experiments with episode length \(T\), we optimize and estimate its truncated version $\widehat J_i^E(\pi_A)
=
\sum_{t=0}^{T-1}
\gamma^t E_i^H(S_t;\pi_A)$,
with empowerment evaluated at each timestep. The assistant receives $E_U^H(S_t;\pi_A)$ as a per-timestep training signal and maximizes the sampled discounted sum over finite rollouts.

Since the assistant objective is to help the user, it is trained to maximize the user's rollout-averaged empowerment, $\pi_A^\star \in \arg\max_{\pi_A} J_U^E(\pi_A)$.
The bystander's empowerment $J_B^E(\pi_A)$ is not part of the assistant's objective; it is evaluated as an externality of the learned assistant policy.

\paragraph{User-Directed Assistance.}
This setting is asymmetric: the assistant is the only learned policy, and it is optimized for the user objective \(J_U^E\). The bystander follows a fixed response model \(\pi_B\), and \(J_B^E(\pi_A)\) is evaluated as an externality of the assistant policy rather than optimized by the assistant. Thus, our setting is not an equilibrium analysis of a Markov game, where each agent strategically best-responds to the others, nor a coalition objective that explicitly aggregates user and bystander empowerment, such as \(J_U^E(\pi_A)+\lambda J_B^E(\pi_A)\). We study the asymmetric case where an assistant is aligned with one target user but deployed in an environment shared with others.

\begin{definition}[Bystander Disempowerment]

Given a learned assistant policy $\pi_A$ and a reference assistant policy
$\pi_A^{\rm ref}$, the bystander disempowerment caused by $\pi_A$ is

\begin{equation}
    D_B(\pi_A,\pi_A^{\rm ref})= J_B^E(\pi_A^{\rm ref})-J_B^E(\pi_A).
\end{equation}

We say that \(\pi_A\) disempowers the bystander relative to \(\pi_A^{\rm ref}\) if $D_B(\pi_A,\pi_A^{\rm ref})>0$.

\end{definition}

 This definition isolates the assistant's contribution to the bystander's change in agency by comparing against a reference assistant operating in the same environment with the same user and bystander response policies.

\input{condition_for_disempowerment}

\begin{figure}[t!]
    \centering
    \includegraphics[width=0.7\linewidth]{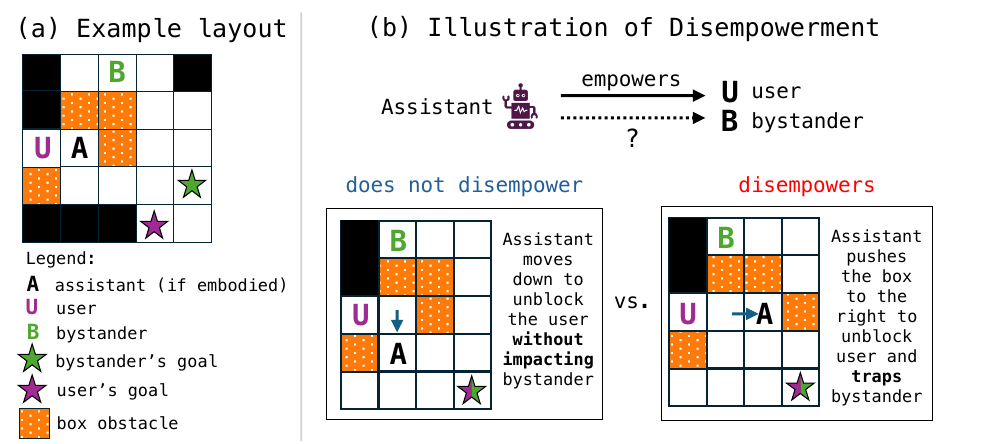}
    \caption{\bench test suite example layout (left) and an illustration of one mode of disempowerment in this environment (right).}
    \label{fig:env_mechanics}
\end{figure}

\section{\bench Test Suite}

We introduce \textbf{\bench}, 
a controlled multi-agent gridworld test suite for isolating and measuring bystander disempowerment. The purpose of the suite is not to model any single deployment domain, but to provide a systematically varied setting in which assistant capabilities, environment dynamics and structure, and assistant objectives can be independently controlled.
\bench is built in JaxMARL for highly efficient training \citep{rutherford_jaxmarl_2024} and will be open sourced to enable further research on disempowerment in multi-agent settings. The environment designs were inspired by prior work \citep{du_ave_2020, leike_ai_2017}, but extend these settings in two key ways: the inclusion of a bystander agent who is not the target of assistance, and goal respawning to enable continuous goal achievement by agents. 

\paragraph{Environment design.}
Each \bench environment is a gridworld containing three agents: a user, a bystander, and an assistant whose objective is to help the user. The environment is parameterized by grid size, number of box obstacles, and number of walls, allowing systematic variation across a combinatorially large configuration space. The environment can also be set up with diverse environmental dynamics, depending on the assistant embodiment and whether direct intervention is enabled (see Figure \ref{fig:env_mechanics}). 

The assistant is capable of pushing or pulling box obstacles, if it is embodied, or of moving any box, if it is non-embodied. If a box is in the path of a user or bystander, they are unable to move through that position. The user and bystander agents pursue their respective goal position in the grid; when an agent reaches its goal, they receive +1 reward, and their goal randomly respawns at another open position. See Appendix \ref{section:disempower_grid_details} for full details.

\paragraph{Disempowerment implementation.} The \bench test suite implements the measurement of each agent's empowerment within and across the training of the assistant. The empowerment measurement is calculated through a Monte-Carlo-based estimation, which samples different rollouts in the gridworld environment (see Appendix \ref{alg:appendix} for empowerment estimation algorithm). The rollout is parameterized by horizon H; in all of our experiments, we set $H=3$. The no assistance reference policy empirically chosen to calculate disempowerment in the experiments is a uniform random policy.

\paragraph{Metrics and training.}
\bench also provides implementations of RL training of the assistant using the goal-agnostic objective of Empowerment Maximization (described in Section \ref{sec:prelim}) and a Goal Inference objective, where the assistant infers and acts on the user's goals \citep{bakerGoalInferenceInverse2007}. For each objective, the assistant receives the corresponding user-centered score as a per-timestep reward and is trained with PPO over finite rollouts. The
bystander's empowerment is never included in the assistant's training objective;
it is only evaluated after training as an externality of the learned assistant
policy. The user and bystander policies are modeled as A* path planning agents during assistant training. See Appendix \ref{subsection:training_details} for more details.

\paragraph{Procedural generation.} \bench supports random procedural generation of environment layouts, with number of walls, wall configurations, number of obstacles, initial obstacle placements, initial goal placements, and initial agent starting positions being varied within a 5×5 grid, spanning a combinatorially large configuration space. See Appendix \ref{alg:appendix} for procedural generation algorithm. We utilize this procedural generation capability in our results to demonstrate that disempowerment is a general consequence that emerges from an assistant's learned policy in shared environments. 

\begin{table}[h]
\centering
\caption{Disempowerment rate (proportion of environments in which the assistant 
disempowers the user) across procedurally generated \bench environments, 
for different assistant capabilities and objectives. Values are rate $\pm$ SE.}
\label{tab:disempowerment_rates}
\begin{tabular}{lcc}
\toprule
Assistant Capability (Least to Most) & Empowerment Maximization & Goal Inference \\
\midrule
Embodied Push-Only (n=100)              & $43\% \pm 5.0\%$ & $49\% \pm 5.0\%$ \\
Embodied (n=100)                        & $56\% \pm 5.0\%$ & $61\% \pm 4.9\%$ \\
Non-Embodied (n=100)                    & $31\% \pm 4.6\%$ & $58\% \pm 4.9\%$ \\
Non-Embodied + Direct Intervention (n=100) & $27\% \pm 4.4\%$ & $96\% \pm 2.0\%$ \\
\bottomrule
\end{tabular}
\label{table:procedural_gen}
\end{table}

\section{Experiments}
\label{sec:generalization}
We procedurally generated $n=400$ random layouts across four different conditions ($n=100$ each), varying in assistant capability, and evaluated bystander disempowerment in each, across two assistant objectives. In the \texttt{Embodied Push-Only} condition, the assistant is embodied and can only push boxes adjacent to its position. In the \texttt{Embodied} condition, the assistant can push or pull boxes adjacent to its position. In the \texttt{Non-Embodied} condition, the assistant is non-embodied and can move any box at any time. In the \texttt{Non-Embodied + Direct Intervention} condition, the assistant is able to not only move any box at any time, but also freeze the bystander for three timesteps at any time.

Between 27\% and 96\% of the generated settings result in bystander disempowerment, demonstrating that disempowerment is a general phenomenon beyond specific initial layouts or environmental dynamics (see Table \ref{tab:disempowerment_rates}). We further analyze how assistant objective and capability impact the presence of disempowerment.

\paragraph{The rate of disempowerment significantly depends on assistance objective when assistant capability is high.}

The rate of disempowerment varies dramatically across two assistance objectives even when the layouts are held fixed, especially when the assistant capability is high. Within-row chi-square tests confirm that the difference between objectives is non-significant for the \texttt{Embodied Push-Only} condition ($\chi^2(1) = 0.72, p = .39$) and \texttt{Embodied} condition ($\chi^2(1) = 0.51, p = .47$) but highly significant for \texttt{Non-Embodied} ($\chi^2(1) = 14.76, p < .01$) and \texttt{Non-Embodied + Direct Intervention} ($\chi^2(1) = 100.54, p < .01$). This provides direct evidence that disempowerment is policy-dependent rather than a static structural property of the environment. This connects directly to Theorem~\ref{thm:local-tradeoff}. The theorem predicts that disempowerment depends on how often the assistant selects actions that help the user at the bystander's expense, not on whether such opportunities exist in the environment. Since the initial layout is fixed, the observed difference across objectives is because of the assistant's policy. When the assistant is more capable, it is better at exploiting whatever opportunities the environment affords, making the choice of objective matter more.

\paragraph{The disempowerment rate of adding direct intervention depends on the assistance objective.}
Adding the ability to freeze the bystander had strikingly different effects depending on the assistant's objective. Under Goal Inference, disempowerment rose to 96\% of generated environments. We hypothesize this is because the assistant treats the bystander's movements as noise interfering with the inferred user goal, and freezing is an effective way to eliminate that noise. Under Empowerment Maximization, however, disempowerment dropped to 27\%. Because user empowerment is computed by rolling out future states assuming a randomly moving bystander, the assistant already anticipates the bystander's movements rather than treating them as interference, which removes the incentive to freeze them in the first place.

\paragraph{Greater capability does not monotonically increase disempowerment.}
Disempowerment rate does not simply increase with assistant capability. Under both assistant objectives, disempowerment insignificantly increases from \texttt{Embodied Push-Only} to \texttt{Embodied} (43\% to 56\% and 49\% to 61\%). However, under Empowerment Maximization, moving from \texttt{Embodied} to \texttt{Non-Embodied} significantly decreases disempowerment from 56\% to 31\%, while under Goal Inference, it insignificantly decreases from 61\% to 58\%. We hypothesize that the larger decrease under Empowerment Maximization is because a non-embodied assistant has more ways to empower the user, since it can move any box freely, rather than being constrained to boxes adjacent to its position, making it less likely to unintentionally disempower the bystander in the process. Under Goal Inference, this additional flexibility provides little benefit to the bystander since the assistant is already focused on the user's goal regardless of its physical constraints. This motivates a closer investigation of specific layouts to understand when and why disempowerment occurs.

\begin{figure}[!t]
    \centering
    \includegraphics[width=1.0\linewidth]{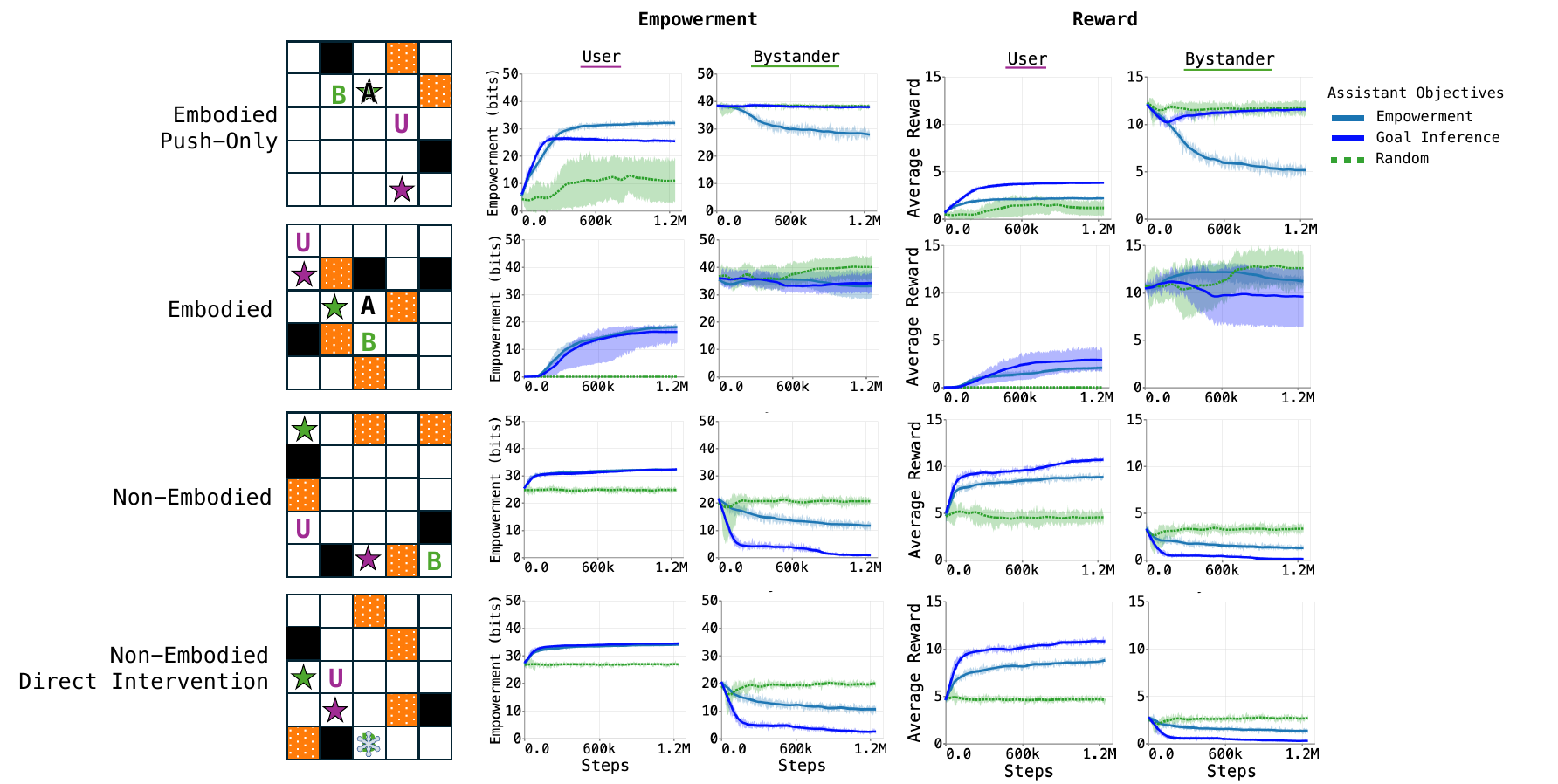}
    \caption{
    Each row corresponds to a sampled layout in each of the four procedurally generated conditions (labeled on left). The layout diagram corresponds to different mechanisms of disempowerment observed in this environment (blocking with embodiment, blocking with boxes, bystander starting in corner, freezing). Center/Right: User and bystander empowerment (1st and 2nd column) and user and bystander reward (3rd and 4th column) across assistant training. Disempowerment is evidenced by higher bystander empowerment under the random policy (green dotted line) than under either assistant objective (blue lines). Each trace is averaged over five runs, with error bands showing standard deviation.}
    \label{fig:stacked_figure}
\end{figure}

\paragraph{Disempowerment emerges through qualitatively distinct mechanisms across conditions.}
To more deeply understand how the assistant disempowers the bystander in \bench, we qualitatively analyze sampled rollouts from the highest-disempowerment layouts in each of the four conditions (see Figure \ref{fig:stacked_figure}).

In the \texttt{Embodied Push-Only} condition, the assistant pushes boxes toward the bystander's goal and, being unable to pull them back, stations itself on top of the bystander's goal for the remainder of the episode. The bystander repeatedly attempts to reach its goal but is blocked. Notably, this mechanism of disempowerment arises under Empowerment Maximization but not Goal Inference, providing further evidence that disempowerment is policy-dependent. A goal-directed assistant is less likely to incidentally occupy the bystander's goal in this way, as it would be actively choosing actions to enable the user to get to their inferred goal.

In the \texttt{Embodied} condition, the assistant rearranges boxes from initially open positions along the sides of the grid to create a narrow constriction, positioning itself within it. Because the assistant has no knowledge of the bystander's goal, it remains in place until the user begins moving in that direction. The high variance across runs suggests that this bottleneck configuration does not emerge consistently across all policy seeds, pointing to some stochasticity in this observed mechanism of disempowerment.

In the \texttt{Non-Embodied} condition, the bystander begins trapped in a corner. The assistant learns to leave the trapping box in place, since the user's goal never spawns at the bystander's location, making it never beneficial to free the bystander. Disempowerment here is thus one of omission rather than active interference. The Goal Inference objective produces more severe disempowerment in this case, as it reasons directly about the user's goal and has no reason to consider the corner; under Empowerment Maximization, the assistant optimizes for the user's control over future states, which includes that region of the grid.

In the \texttt{Non-Embodied + Direct Intervention} condition, the assistant repeatedly freezes the bystander whenever it attempts to move while the user is nearby. This represents a qualitatively distinct mode of disempowerment. Rather than shaping the environment in which the bystander acts, the assistant directly eliminates the bystander's ability to transition to new states. Under Goal Inference, the bystander's movements are treated as noise interfering with the inferred user goal, and freezing serves as an effective means of elimination. Under Empowerment Maximization, the assistant models the bystander as a random agent when computing future empowerment, allowing it to anticipate rather than suppress the bystander's movements, which reduces the incentive to freeze.

\section{Discussion}


\paragraph{Implications for AI Safety and Cooperative AI.}
Our formalization and experiments demonstrate that a well-meaning personal AI assistant can significantly decrease a bystander's control over its environment. The assistant selected actions that benefited the user at the bystander’s expense, even when reversible or mutually beneficial actions were available. Ideally, we would be able to anticipate and prevent disempowerment. However, predicting disempowerment is a challenge, as it depends on the exact learned assistant objectives and temporal environmental dynamics. As a first step towards this, disempowerment enables an agent-centric measure of harm caused by an AI agent in a multi-agent setting.

\paragraph{Preliminary Mitigation.}
We conducted preliminary experiments with a joint empowerment objective that sums user and bystander empowerment \citep{van2020robust} over 110 generated layouts. We found that this approach substantially mitigates disempowerment, with a tradeoff. In the 106 layouts where an empowerment-maximizing assistant disempowers the bystander, joint empowerment instead increased bystander empowerment in 52\% of cases ($p<0.001$, $d=0.78$) and produced no significant impact in the remaining 48\%. However, this came at a significant cost to user reward ($p<0.001$), revealing a fundamental trade-off between bystander disempowerment and user reward. Joint empowerment utilized equal weights for user and bystander. Choosing appropriate weights is non-trivial, with no ground truth as to who to prioritize. Joint empowerment also faces a scalability limitation in environments with multiple bystanders, as it requires potentially unrealistic assumptions about the assistant's familiarity with each bystander's action space. See Appendix~\ref{paragraph:joint_empowerment} for full details.

\paragraph{Future Work.}
Our work introduces a measure of disempowerment and characterizes how it arises across environment dynamics and assistant capabilities, revealing a central open question: how can assistance objectives be designed to benefit an intended user without unintentionally harming others? One important extension is to consider how disempowerment applies when the bystander is adaptive and responds to repeated disempowerment over time, which could reveal whether disempowerment persists or resolves under different bystander policies \citep{shen2024towards}. A limitation of our work is our usage of discrete gridworld environments, which may not reflect real-world multi-agent dynamics with continuous state spaces, partial observability, and temporally extended interactions; extending disempowerment computation to such settings will require tractable empowerment approximations. Despite this, disempowerment is a valuable metric for measuring agent-centric harm, and by formalizing it and characterizing when and how it occurs, we take a first step toward ensuring that personal AI agents designed to help one person do not inadvertently harm another.

\bibliographystyle{plainnat}
\bibliography{references}


\appendix
\input{appendix_theory_version_1}



\section{\bench Details}
\label{section:disempower_grid_details}
\begin{table}[h]
    \caption{Command-line parameters for environment configuration.}
    \centering
    \begin{small}
    \begin{tabular}{lll}
    \toprule
    \textbf{Parameter} & \textbf{Type} & \textbf{Description} \\
    \midrule
    \texttt{--grid\_height} & int & Height of the grid environment \\
    \texttt{--grid\_width} & int & Width of the grid environment \\
    \texttt{--num\_boxes} & int ($\geq 1$) & Number of boxes in the environment \\
    \texttt{--num\_goals} & int (1--2) & Number of goals in the environment \\
    \texttt{--num\_walls} & int & Number of walls in the environment \\
    \texttt{--max\_steps} & int (default=50) & Maximum number of steps per episode \\
    \texttt{--helper\_objective} & string & Helper objective (e.g., empowerment, random) \\
    \texttt{--goal\_respawn\_seed} & int & Seed for random reproducible agent goal respawn locations \\
    \texttt{--epochs} & int (default=250) & Number of training epochs \\
    \texttt{--specific\_positions\_file} & string & JSON file specifying initial environment layout \\
    \texttt{--no\_freeze} & flag & Disable assistant freezing bystander action \\
    \texttt{--no\_pull} & flag & Disable assistant pulling boxes action (if assistant is embodied) \\
    \texttt{--no\_goal\_respawn} & flag & Disable goal respawning \\
    \bottomrule
    \end{tabular}
    \end{small}
    \label{tab:parameters}
\end{table}

\paragraph{Environment Details.} In \bench, $R_U(s_t)=1$ if the user reaches its assigned goal $g_U \in S$, $0$ otherwise. $R_B(s_t)=1$ if the bystander reaches its assigned goal $g_B \in S$, $0$ otherwise. The user and bystander may be assigned to the same goal or different goals. Regardless, the reward each agent receives is fully independent of that of the other agent. In that case, the state to observation mapping function differs between the human and assistant, if the assistant is maximizing the user's empowerment. $\Omega_H$ includes the goals pursued by the user and bystander, while $\Omega_A$ does not, i.e., the assistant has no knowledge of the user or bystander's goal. If the assistant is maximizing the likelihood of the user reaching its goal, $\Omega_A$ includes knowledge of the user's goal.

At time $t$, the humans (user and bystander) observe $\omega^H_t \in \Omega_H(s_t)$, and the assistant observes $\omega^A_t \in \Omega_A(s_t)$. Action selection happens simultaneously. The user selects action $a^U_t \sim \pi_{U}(\cdot|\omega^H_t)$, the bystander selects action $a^B_t \sim \pi_{B}(\cdot|\omega^H_t)$, and the assistant selects action $a^A_t \sim \pi_A(\cdot | \omega^A_t)$.

\section{Training Details}
\label{subsection:training_details}
$\pi_A$ is trained using PPO. The user and bystander act according to their fixed A* path planning policies $\pi_{U}$ and $\pi_{B}$, respectively. This models an assistant learning its policy while interacting with adaptive humans that act according to their underlying goals.

All experiments were run on a single NVIDIA GeForce RTX 4090 GPU (24 GB VRAM) and are reproducible on CPU, GPU, or TPU. Running PPO training for 250 epochs takes around 3 minutes for one assistant objective in one environment on the 4090 GPU, and would likely take at least 5x more time on CPU, due to loss of JAX speedup.

\begin{table}[!h]
\centering
\caption{\bench experiment configuration}
\label{tab:ppo_env_config}
\begin{small}
\begin{tabular}{ll}
\toprule
\textbf{Parameter} & \textbf{Value} \\
\midrule
Number of parallel environments ($N_{\text{envs}}$) & 100 \\
Steps per episode ($T$) & 50 \\
Total training timesteps & $1.25 \times 10^{6}$ \\
Training epochs & 250 \\
\bottomrule
\end{tabular}
\end{small}
\end{table}

\begin{table}[!h]
\centering
\caption{PPO optimization hyperparameters}
\label{tab:ppo_optim_config}
\begin{small}
\begin{tabular}{ll}
\toprule
\textbf{Parameter} & \textbf{Value} \\
\midrule
Learning rate annealing & Yes \\
Learning rate ($\alpha$) & $3 \times 10^{-4}$ \\
Discount factor ($\gamma$) & 0.99 \\
GAE parameter ($\lambda$) & 0.95 \\
Clipping parameter ($\epsilon$) & 0.2 \\
PPO update epochs ($K$) & 4 \\
Minibatches per update ($M$) & 4 \\
Value function coefficient ($c_v$) & 0.5 \\
Entropy coefficient ($c_e$) & 0.1 \\
Maximum gradient norm & 0.5 \\
Activation function & \texttt{tanh} \\
\bottomrule
\end{tabular}
\end{small}
\end{table}


\section{Code}
We include the code as an anonymized ZIP file in the submission, as \bench is an important contribution of this work and will be open-sourced.

\section{Joint Empowerment}
\label{paragraph:joint_empowerment}
A naive approach to preventing disempowerment is to include the bystander's empowerment in the assistant's objective alongside the user's. Van der Heiden et al. \citep{van2020robust} originally proposed this approach and showed that it improves multi-agent coordination in cooperative tasks. Rather than maximizing only the user's empowerment, the assistant maximizes the sum of both user and bystander agents' individual empowerment:
\begin{equation}
    \pi_A^\star \in \arg\max_{\pi_A} (J_U^E(\pi_A) + J_B^E(\pi_A))
\end{equation}

We evaluate this objective over 110 generated layouts that systematically vary key-goal placements. The key must be picked up by the agent before entering the goal to receive reward. In the 106 layouts where an empowerment-maximizing assistant disempowers the bystander, a joint empowerment assistant increases bystander empowerment in 52\% of cases ($p<0.001$, $d=0.78$) and produces no significant impact in the remaining 48\% ($p=0.10$). However, across all 106 layouts, an assistant maximizing joint empowerment significantly reduced the user's reward ($p<0.001$) while simultaneously and significantly increasing the bystander's reward ($p<0.001$), compared to an assistant optimizing empowerment. 

The user reward cost raises a deeper question: an assistant that accepts reduced user reward to preserve bystander agency may in some sense be better aligned with broader societal values — prioritizing collective welfare over individual task performance. Whether this tradeoff is desirable depends on normative assumptions about whose interests the assistant should serve, and we leave a principled treatment of this for future work.


\begin{figure}[H]
    \centering
    \includegraphics[width=1\linewidth]{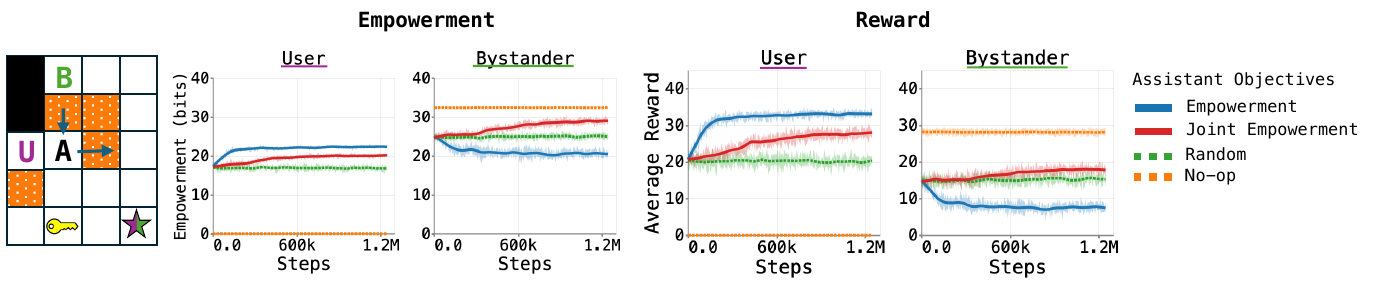}
    \caption{In this layout with the \texttt{Embodied} environmental dynamics and no goal respawn with a key that must be picked up before entering the goal, the assistant maximizing the joint empowerment avoids disempowering the bystander, but significantly decreases the user's empowerment and reward, compared to when maximizing the user's empowerment. Importantly, note that joint empowerment still performs significantly better than an assistant acting randomly for both the user and bystander's empowerment and reward. }
    \label{fig:joint_empowerment_vs_empowerment}
\end{figure}

\clearpage
\section{Algorithms}
\label{alg:appendix}

\begin{algorithm}
\caption{Monte Carlo Empowerment Estimation}
\begin{algorithmic}[1]
\REQUIRE State $s_0$, Horizon $H$, Trajectories $N$, Action set $\mathcal{A}$
\ENSURE Empowerment $\mathcal{E}$ for each agent
\FOR{each agent $i \in \{U, B\}$}
    \FOR{each action $a \in \mathcal{A}$}
        \STATE $\mathcal{P}_{a} \leftarrow \text{zeros}(\text{GridSize})$
        \FOR{$n = 1$ \TO $N$}
            \STATE $s \leftarrow s_0$
            \STATE \COMMENT{First step: Agent $i$ takes action $a$}
            \STATE $a_{i} \leftarrow a$, $a_{others} \sim \text{Uniform}(\mathcal{A})$
            \STATE $s \leftarrow \text{Transition}(s, a_{i}, a_{others})$
            \FOR{$t = 2$ \TO $H$}
                \STATE \COMMENT{Remaining steps: All agents move randomly}
                \STATE $a_{all} \sim \text{Uniform}(\mathcal{A})$
                \STATE $s \leftarrow \text{Transition}(s, a_{all})$
            \ENDFOR
            \STATE $pos \leftarrow \text{GetPos}(s, i)$
            \STATE $\mathcal{P}_{a}[pos] \leftarrow \mathcal{P}_{a}[pos] + \frac{1}{N}$
        \ENDFOR
    \ENDFOR
    \STATE $P(s^+|s_0) \leftarrow \frac{1}{|\mathcal{A}|} \sum_{a} \mathcal{P}_{a}$
    \STATE $\mathcal{E}_i \leftarrow \sum_{a} \frac{1}{|\mathcal{A}|} \sum_{s^+} \mathcal{P}_{a}[s^+] \log \frac{\mathcal{P}_{a}[s^+]}{P(s^+|s_0)}$
\ENDFOR
\RETURN $\mathcal{E}$
\end{algorithmic}
\end{algorithm}

\begin{algorithm}[h]
\caption{Procedural Generation of Random Valid Multi-Agent Gridworld Layouts}
\begin{algorithmic}[1]
\REQUIRE Grid size $(H, W)$, number of boxes $B$, wall density $\rho$, maximum attempts $N$
\ENSURE Layout where (i) user starts trappable, (ii) user and bystander goals are reachable

\FOR{$i = 1$ to $N$}
    \STATE Sample random wall positions with density $\rho$
    \STATE Build environment graph $G$ from walls
    \STATE Identify accessible (non-wall) cells
    \IF{insufficient accessible cells}
        \STATE \textbf{continue}
    \ENDIF

    \STATE Identify trappable user start positions (few accessible neighbors)
    \IF{no trappable positions found}
        \STATE \textbf{continue}
    \ENDIF

    \FOR{multiple placement attempts}
        \STATE Select trappable position for user
        \STATE Reserve adjacent cells for trapping boxes

        \STATE Select user goal reachable from user (ignoring boxes)
        \IF{no reachable goal}
            \STATE \textbf{continue}
        \ENDIF

        \STATE Select bystander start position
        \STATE Select bystander goal reachable from bystander
        \IF{no reachable bystander goal}
            \STATE \textbf{continue}
        \ENDIF

        \STATE Place trapping boxes adjacent to user
        \STATE Compute bystander’s shortest path to goal
        \STATE Place remaining boxes away from bystander path

        \IF{bystander goal becomes unreachable}
            \STATE \textbf{continue}
        \ENDIF

        \STATE \textbf{return} valid layout
    \ENDFOR
\ENDFOR

\STATE \textbf{return} failure
\end{algorithmic}
\end{algorithm}



\end{document}

%% file: condition_for_disempowerment.tex
\section{Conditions for Bystander Disempowerment}

We now characterize how a user-directed assistant can reduce a bystander's empowerment. The key mechanism 
is that disempowerment arises when the assistant's objective shifts probability toward actions that increase the user's current and future empowerment while decreasing the bystander's current and future empowerment. We first state this mechanism abstractly, then show that assistant-controllable bottlenecks are a concrete structural special case.

\subsection{Local Tradeoffs in Empowerment Value}
The rollout-averaged empowerment objective \(J_i^E(\pi_A)\) from Section~\ref{sec:prelim} induces standard value-function analogues. For \(i\in\{U,B\}\), define the empowerment value of state \(s\) under assistant policy \(\pi_A\) as
\[
V_i^E(s;\pi_A)
=
\mathbb{E}_{\pi_A,\pi_U,\pi_B}
\left[
\sum_{\tau=0}^{\infty}
\gamma^\tau E_i^H(S_\tau;\pi_A)
\;\middle|\;
S_0=s
\right],
\]
and the state-action empowerment value as
\[
Q_i^E(s,a_A;\pi_A)
=
\mathbb{E}_{\pi_A,\pi_U,\pi_B}
\left[
\sum_{\tau=0}^{\infty}
\gamma^\tau E_i^H(S_\tau;\pi_A)
\;\middle|\;
S_0=s,\ A_0^A=a_A
\right],
\]
where after the first assistant action \(a_A\), the assistant follows \(\pi_A\). 

\begin{definition}[Local Tradeoff Action]
Fix a learned assistant policy \(\pi_A\) and a reference assistant
policy \(\pi_A^{\rm ref}\). For an assistant action \(a_A\) at state
\(s\), define the local user gain
\[
g_U(s,a_A)
=
Q_U^E(s,a_A;\pi_A)
-
\mathbb{E}_{a\sim \pi_A^{\rm ref}(\cdot\mid s)}
Q_U^E(s,a;\pi_A^{\rm ref}),
\]
and the local bystander loss
\[
\ell_B(s,a_A)
=
\mathbb{E}_{a\sim \pi_A^{\rm ref}(\cdot\mid s)}
Q_B^E(s,a;\pi_A^{\rm ref})
-
Q_B^E(s,a_A;\pi_A).
\]
An action \(a_A\) is a \((\delta_U,\delta_B)\)-local tradeoff action at
state \(s\) if
\[
g_U(s,a_A)\ge \delta_U>0
\quad\text{and}\quad
\ell_B(s,a_A)\ge \delta_B>0.
\]
Let \(\mathcal{H}_{\delta_U,\delta_B}(s)\) denote the set of such
actions.
\end{definition}

A local tradeoff action is therefore one that improves the user's empowerment value relative to the reference assistant while decreasing the bystander's empowerment value. The definition is policy-level: it does not require a particular training algorithm. Some assistant actions may benefit both the user and the bystander, or harm neither. The question is whether the learned assistant selects tradeoff actions often enough, over the states it visits, for the bystander's net empowerment to decrease.

\begin{definition}[Systematic Tradeoff Mass]

Let \(d_{\pi_A}\) be the discounted state occupancy measure induced by
\((\pi_A,\pi_U,\pi_B)\): $d_{\pi_A}(s)
=
\mathbb{E}_{\pi_A,\pi_U,\pi_B}
\left[
\sum_{t=0}^{\infty}
\gamma^t \mathbf{1}\{S_t=s\}
\right]$.
For a learned assistant \(\pi_A\), define its cumulative excess tradeoff mass relative to \(\pi_A^{\rm ref}\) as
\[
\eta(\pi_A,\pi_A^{\rm ref})
=
\sum_s d_{\pi_A}(s)
\left[
\pi_A(\mathcal{H}_{\delta_U,\delta_B}(s)\mid s)
-
\pi_A^{\rm ref}(\mathcal{H}_{\delta_U,\delta_B}(s)\mid s)
\right]_+.
\]
\end{definition}

\subsection{Systematic Local Tradeoffs Imply Disempowerment}

\begin{theorem}[Systematic local tradeoffs imply bystander disempowerment]\label{thm:local-tradeoff}
Let \(\pi_A\) be a learned assistant and \(\pi_A^{\rm ref}\) a reference assistant. Suppose \(\pi_A\) has cumulative excess tradeoff mass \(\eta(\pi_A,\pi_A^{\rm ref})\ge \eta\) for \((\delta_U,\delta_B)\)-local tradeoff actions. Suppose further that any increase in bystander empowerment value outside these tradeoff state-action pairs is at most \(\epsilon_{\rm out}\). If $\eta\delta_B>\epsilon_{\rm out}$,
then $J_B^E(\pi_A)<J_B^E(\pi_A^{\rm ref})$.
That is, $D_B(\pi_A,\pi_A^{\rm ref})>0$. Thus, \(\pi_A\) disempowers the bystander relative to
\(\pi_A^{\rm ref}\).
\end{theorem}

\begin{proof}[Proof sketch]
Each selected local tradeoff action decreases the bystander's empowerment value by at least \(\delta_B\) relative to the reference assistant. Since the learned assistant selects such actions with cumulative excess mass at least \(\eta\) 
, these actions contribute total bystander empowerment loss at least \(\eta\delta_B\). By assumption, all bystander gains outside the tradeoff state-action pairs are bounded by \(\epsilon_{\rm out}\). When \(\eta\delta_B>\epsilon_{\rm out}\), the losses dominate the gains, so
\(J_B^E(\pi_A)<J_B^E(\pi_A^{\rm ref})\). We provide full proof in Appendix~\ref{appsec:proof_of_theorem}.
\end{proof}

%% file: appendix_theory_version_1.tex
\section{Proofs}
\label{app:proofs}


\subsection{Proof of Theorem~\ref{thm:local-tradeoff}}\label{appsec:proof_of_theorem}

We first restate the relevant definitions. For \(i\in\{U,B\}\), define
the empowerment value
\[
V_i^E(s;\pi_A)
=
\mathbb{E}_{\pi_A,\pi_U,\pi_B}
\left[
\sum_{\tau=0}^{\infty}
\gamma^\tau E_i^H(S_\tau;\pi_A)
\;\middle|\;
S_0=s
\right],
\]
and the state-action empowerment value
\[
Q_i^E(s,a_A;\pi_A)
=
\mathbb{E}_{\pi_A,\pi_U,\pi_B}
\left[
\sum_{\tau=0}^{\infty}
\gamma^\tau E_i^H(S_\tau;\pi_A)
\;\middle|\;
S_0=s,\ A_0^A=a_A
\right].
\]
The rollout-averaged bystander empowerment is
$J_B^E(\pi_A)
=
\mathbb{E}_{S_0\sim \rho_0}
\left[
V_B^E(S_0;\pi_A)
\right]$,
where \(\rho_0\) is the initial-state distribution.

Fix a learned assistant policy \(\pi_A\) and a reference assistant policy
\(\pi_A^{\rm ref}\). For an assistant action \(a_A\) at state \(s\),
define the local bystander loss
\[
\ell_B(s,a_A)
=
\mathbb{E}_{a\sim \pi_A^{\rm ref}(\cdot\mid s)}
(Q_B^E(s,a;\pi_A^{\rm ref})
-
Q_B^E(s,a_A;\pi_A)).
\]
Let \(\mathcal{H}_{\delta_U,\delta_B}(s)\) be the set of
\((\delta_U,\delta_B)\)-local tradeoff actions at state \(s\). By
definition, if \(a_A\in \mathcal{H}_{\delta_U,\delta_B}(s)\), then $\ell_B(s,a_A)\ge \delta_B$.

Let \(d_{\pi_A}\) be the discounted state occupancy measure induced by
\((\pi_A,\pi_U,\pi_B)\):
\[
d_{\pi_A}(s)
=
\mathbb{E}_{\pi_A,\pi_U,\pi_B}
\left[
\sum_{t=0}^{\infty}
\gamma^t \mathbf{1}\{S_t=s\}
\right].
\]
The cumulative excess tradeoff mass is
\[
\eta(\pi_A,\pi_A^{\rm ref})
=
\sum_s d_{\pi_A}(s)
\left[
\pi_A(\mathcal{H}_{\delta_U,\delta_B}(s)\mid s)
-
\pi_A^{\rm ref}(\mathcal{H}_{\delta_U,\delta_B}(s)\mid s)
\right]_+.
\]

We now decompose the bystander empowerment difference
$D_B(\pi_A,\pi_A^{\rm ref})
=
J_B^E(\pi_A^{\rm ref})-J_B^E(\pi_A)$
into the contribution from local tradeoff state-action pairs and the
remaining contribution. Let $\mathcal{T}
=
\{(s,a_A):a_A\in \mathcal{H}_{\delta_U,\delta_B}(s)\}$
denote the set of local tradeoff state-action pairs.

By assumption, \(\pi_A\) selects local tradeoff actions with cumulative
excess mass at least \(\eta\):$\eta(\pi_A,\pi_A^{\rm ref})\ge \eta$.
Each such local tradeoff action decreases the bystander's empowerment
value by at least \(\delta_B\) relative to the reference assistant.
Therefore, the total bystander empowerment loss contributed by excess
selection of tradeoff actions is at least $\eta\delta_B$.

Let \(\epsilon_{\rm out}\) upper bound all offsetting effects outside
these excess local tradeoff selections. Write
$D_B(\pi_A,\pi_A^{\rm ref})
=
L_{\mathcal{T}} - G_{\rm out}$,
where \(L_{\mathcal{T}}\) is the bystander empowerment loss contributed
by excess selection of local tradeoff actions and \(G_{\rm out}\) is the
net bystander empowerment gain, if any, from all other differences
between \(\pi_A\) and \(\pi_A^{\rm ref}\), including non-tradeoff
actions, different state occupancies, and possible beneficial side
effects of the learned assistant. The bounded-offset assumption states
that $G_{\rm out}\le \epsilon_{\rm out}$.
Since \(L_{\mathcal{T}}\ge \eta\delta_B\), we obtain
$D_B(\pi_A,\pi_A^{\rm ref})
=
L_{\mathcal{T}}-G_{\rm out}
\ge
\eta\delta_B-\epsilon_{\rm out}$.
If $\eta\delta_B>\epsilon_{\rm out}$,
then $D_B(\pi_A,\pi_A^{\rm ref})>0$.
Equivalently, $J_B^E(\pi_A)<J_B^E(\pi_A^{\rm ref})$.
Thus, \(\pi_A\) disempowers the bystander relative to the reference
assistant.

\subsection{Assistant-controllable bottlenecks as a structural special case} \label{appsec:bottleneck}
\begin{definition}[Empowerment-reducing bottleneck closure]
At state \(s_c\), an assistant action \(a^{\rm close}\) is an
empowerment-reducing bottleneck closure relative to \(a^{\rm open}\) if
\[
\mathcal{R}_B^H(s_c,a^{\rm close})
\subsetneq
\mathcal{R}_B^H(s_c,a^{\rm open})
\]
and
\[
Q_B^E(s_c,a^{\rm open};\pi_A)
-
Q_B^E(s_c,a^{\rm close};\pi_A)
\ge \delta_B
\]
for some \(\delta_B>0\).
\end{definition}

\begin{proposition}[Bottleneck closure induces a local tradeoff]
\label{prop:bottleneck-local-tradeoff}
Suppose \(a^{\rm close}\) is an empowerment-reducing bottleneck closure
relative to \(a^{\rm open}\) at state \(s_c\). If
\[
Q_U^E(s_c,a^{\rm close};\pi_A)
-
Q_U^E(s_c,a^{\rm open};\pi_A)
\ge \delta_U
\]
for some \(\delta_U>0\), then \(a^{\rm close}\) is a
\((\delta_U,\delta_B)\)-local tradeoff action.
\end{proposition}

\begin{proof}
By the definition of empowerment-reducing bottleneck closure,
\[
Q_B^E(s_c,a^{\rm open};\pi_A)
-
Q_B^E(s_c,a^{\rm close};\pi_A)
\ge \delta_B.
\]
By assumption,
\[
Q_U^E(s_c,a^{\rm close};\pi_A)
-
Q_U^E(s_c,a^{\rm open};\pi_A)
\ge \delta_U.
\]
Thus closing the bottleneck increases the user's empowerment value by at
least \(\delta_U\) while decreasing the bystander's empowerment value by
at least \(\delta_B\), so it is a \((\delta_U,\delta_B)\)-local tradeoff
action.
\end{proof}
\begin{corollary}[Systematic bottleneck closure implies bystander disempowerment]
\label{cor:bottleneck-disempowerment}
Suppose that the learned assistant selects empowerment-reducing
bottleneck closures with cumulative excess mass at least \(\eta\)
relative to \(\pi_A^{\rm ref}\). Suppose each closure reduces the
bystander's empowerment value by at least \(\delta_B\), and suppose that
all offsetting bystander gains outside these closures are bounded by
\(\epsilon_{\rm out}\). If
\[
\eta\delta_B>\epsilon_{\rm out},
\]
then
\[
D_B(\pi_A,\pi_A^{\rm ref})>0.
\]
Thus, the learned assistant disempowers the bystander.
\end{corollary}

\begin{proof}
By Proposition~\ref{prop:bottleneck-local-tradeoff}, each such closure
is a local tradeoff action. The learned assistant selects these actions
with cumulative excess mass at least \(\eta\), and each produces
bystander empowerment loss at least \(\delta_B\). Therefore the total
loss from systematic bottleneck closure is at least \(\eta\delta_B\).
Since all offsetting gains are bounded by \(\epsilon_{\rm out}\), the
condition \(\eta\delta_B>\epsilon_{\rm out}\) implies by
Theorem~\ref{thm:local-tradeoff} that
\(D_B(\pi_A,\pi_A^{\rm ref})>0\).
\end{proof}

%% file: references.bib
@misc{turner_optimal_2023,
	title = {Optimal {Policies} {Tend} to {Seek} {Power}},
	url = {http://arxiv.org/abs/1912.01683},
	doi = {10.48550/arXiv.1912.01683},
	urldate = {2024-10-08},
	publisher = {arXiv},
	author = {Turner, Alexander Matt and Smith, Logan and Shah, Rohin and Critch, Andrew and Tadepalli, Prasad},
	month = jan,
	year = {2023},
	note = {arXiv:1912.01683 [cs]},
}

@inproceedings{franzmeyer_learning_2022,
	title = {Learning {Altruistic} {Behaviours} in {Reinforcement} {Learning} without {External} {Rewards}},
	url = {https://openreview.net/forum?id=KxbhdyiPHE},
	booktitle = {International {Conference} on {Learning} {Representations}},
	author = {Franzmeyer, Tim and Malinowski, Mateusz and Henriques, Joao F.},
	year = {2022},
}

@inproceedings{myers_learning_2024,
	title = {Learning to {Assist} {Humans} without {Inferring} {Rewards}},
	volume = {37},
	booktitle = {Advances in {Neural} {Information} {Processing} {Systems}},
	publisher = {Curran Associates, Inc.},
	author = {Myers, Vivek and Ellis, Evan and Levine, Sergey and Eysenbach, Benjamin and Dragan, Anca},
	editor = {Globerson, A. and Mackey, L. and Belgrave, D. and Fan, A. and Paquet, U. and Tomczak, J. and Zhang, C.},
	year = {2024},
	pages = {71540--71567},
}

@inproceedings{krakovna_avoiding_2020,
	address = {Red Hook, NY, USA},
	series = {{NIPS} '20},
	title = {Avoiding {Side Effects} by {Considering Future Tasks}},
	isbn = {978-1-7138-2954-6},
	urldate = {2025-09-14},
	booktitle = {Proceedings of the 34th {International} {Conference} on {Neural} {Information} {Processing} {Systems}},
	publisher = {Curran Associates Inc.},
	author = {Krakovna, Victoria and Orseau, Laurent and Ngo, Richard and Martic, Miljan and Legg, Shane},
	month = dec,
	year = {2020},
	pages = {19064--19074},
}

@inproceedings{du_ave_2020,
	title = {{AvE}: {Assistance} via {Empowerment}},
	volume = {33},
	shorttitle = {{AvE}},
	booktitle = {Advances in {Neural} {Information} {Processing} {Systems}},
	publisher = {Curran Associates, Inc.},
	author = {Du, Yuqing and Tiomkin, Stas and Kiciman, Emre and Polani, Daniel and Abbeel, Pieter and Dragan, Anca},
	year = {2020},
	pages = {4560--4571},
}

@inproceedings{hadfield-menell_cooperative_2016,
	address = {Red Hook, NY, USA},
	series = {{NIPS}'16},
	title = {{Cooperative} {Inverse} {Reinforcement} {Learning}},
	isbn = {978-1-5108-3881-9},
	urldate = {2025-09-14},
	booktitle = {Proceedings of the 30th {International} {Conference} on {Neural} {Information} {Processing} {Systems}},
	publisher = {Curran Associates Inc.},
	author = {Hadfield-Menell, Dylan and Dragan, Anca and Abbeel, Pieter and Russell, Stuart},
	month = dec,
	year = {2016},
	pages = {3916--3924},
}

@misc{amodei_concrete_2016,
	title = {Concrete {Problems} in {AI} {Safety}},
	url = {http://arxiv.org/abs/1606.06565},
	doi = {10.48550/arXiv.1606.06565},
	language = {en},
	urldate = {2025-09-14},
	publisher = {arXiv},
	author = {Amodei, Dario and Olah, Chris and Steinhardt, Jacob and Christiano, Paul and Schulman, John and Mané, Dan},
	month = jul,
	year = {2016},
	note = {arXiv:1606.06565 [cs]},
}

@misc{rutherford_jaxmarl_2024,
	title = {{JaxMARL}: {Multi}-{Agent} {RL} {Environments} and {Algorithms} in {JAX}},
	shorttitle = {{JaxMARL}},
	url = {http://arxiv.org/abs/2311.10090},
	doi = {10.48550/arXiv.2311.10090},
	language = {en},
	urldate = {2025-08-12},
	publisher = {arXiv},
	author = {Rutherford, Alexander and Ellis, Benjamin and Gallici, Matteo and Cook, Jonathan and Lupu, Andrei and Ingvarsson, Gardar and Willi, Timon and Hammond, Ravi and Khan, Akbir and Witt, Christian Schroeder de and Souly, Alexandra and Bandyopadhyay, Saptarashmi and Samvelyan, Mikayel and Jiang, Minqi and Lange, Robert Tjarko and Whiteson, Shimon and Lacerda, Bruno and Hawes, Nick and Rocktaschel, Tim and Lu, Chris and Foerster, Jakob Nicolaus},
	month = nov,
	year = {2024},
	note = {arXiv:2311.10090 [cs]},
}

@misc{leike_ai_2017,
	title = {{AI} {Safety} {Gridworlds}},
	url = {http://arxiv.org/abs/1711.09883},
	doi = {10.48550/arXiv.1711.09883},
	language = {en},
	urldate = {2025-08-11},
	publisher = {arXiv},
	author = {Leike, Jan and Martic, Miljan and Krakovna, Victoria and Ortega, Pedro A. and Everitt, Tom and Lefrancq, Andrew and Orseau, Laurent and Legg, Shane},
	month = nov,
	year = {2017},
	note = {arXiv:1711.09883 [cs]},
}

@inproceedings{klyubin_empowerment_2005,
	title = {Empowerment: a {Universal Agent-Centric Measure} of {Control}},
	volume = {1},
	shorttitle = {Empowerment},
	url = {https://ieeexplore.ieee.org/document/1554676/},
	doi = {10.1109/CEC.2005.1554676},
	urldate = {2025-05-27},
	booktitle = {2005 {IEEE} {Congress} on {Evolutionary} {Computation}},
	author = {Klyubin, A.S. and Polani, D. and Nehaniv, C.L.},
	month = sep,
	year = {2005},
	note = {ISSN: 1941-0026},
	pages = {128--135 Vol.1},
}

@misc{jung_empowerment_2012,
	title = {Empowerment for {Continuous} {Agent}-{Environment} {Systems}},
	url = {http://arxiv.org/abs/1201.6583},
	doi = {10.48550/arXiv.1201.6583},
	language = {en},
	urldate = {2025-05-27},
	publisher = {arXiv},
	author = {Jung, Tobias and Polani, Daniel and Stone, Peter},
	month = jan,
	year = {2012},
	note = {arXiv:1201.6583 [cs]},
}

@inproceedings{klyubin_all_2005,
	address = {Berlin, Heidelberg},
	title = {All {Else} {Being} {Equal} {Be} {Empowered}},
	isbn = {978-3-540-31816-3},
	doi = {10.1007/11553090_75},
	language = {en},
	booktitle = {Advances in {Artificial} {Life}},
	publisher = {Springer},
	author = {Klyubin, Alexander S. and Polani, Daniel and Nehaniv, Chrystopher L.},
	editor = {Capcarrère, Mathieu S. and Freitas, Alex A. and Bentley, Peter J. and Johnson, Colin G. and Timmis, Jon},
	year = {2005},
	pages = {744--753},
}

@article{brandle_empowerment_2023,
	title = {Empowerment {Contributes} to {Exploration Behaviour} in a {Creative Video Game}},
	volume = {7},
	copyright = {2023 The Author(s), under exclusive licence to Springer Nature Limited},
	issn = {2397-3374},
	url = {https://www.nature.com/articles/s41562-023-01661-2},
	doi = {10.1038/s41562-023-01661-2},
	language = {en},
	number = {9},
	urldate = {2025-05-13},
	journal = {Nature Human Behaviour},
	author = {Brändle, Franziska and Stocks, Lena J. and Tenenbaum, Joshua B. and Gershman, Samuel J. and Schulz, Eric},
	month = sep,
	year = {2023},
	note = {Publisher: Nature Publishing Group},
	pages = {1481--1489},
}

@misc{lidayan_intrinsically-motivated_2025,
	title = {Intrinsically-{Motivated} {Humans} and {Agents} in {Open}-{World} {Exploration}},
	url = {http://arxiv.org/abs/2503.23631},
	doi = {10.48550/arXiv.2503.23631},
	language = {en},
	urldate = {2025-04-17},
	publisher = {arXiv},
	author = {Lidayan, Aly and Du, Yuqing and Kosoy, Eliza and Rufova, Maria and Abbeel, Pieter and Gopnik, Alison},
	month = mar,
	year = {2025},
	note = {arXiv:2503.23631 [cs]},
}

@misc{baddam_search_2025,
	title = {In {Search} of a {Lost} {Metric}: {Human} {Empowerment} as a {Pillar} of {Socially} {Conscious} {Navigation}},
	shorttitle = {In {Search} of a {Lost} {Metric}},
	url = {http://arxiv.org/abs/2501.01539},
	doi = {10.48550/arXiv.2501.01539},
	language = {en},
	urldate = {2025-02-06},
	publisher = {arXiv},
	author = {Baddam, Vasanth Reddy and Chalaki, Behdad and Tadiparthi, Vaishnav and Mahjoub, Hossein Nourkhiz and Moradi-Pari, Ehsan and Eldardiry, Hoda and Boker, Almuatazbellah},
	month = jan,
	year = {2025},
	note = {arXiv:2501.01539 [cs]},
}

@inproceedings{turner_conservative_2020,
	address = {New York, NY, USA},
	series = {{AIES} '20},
	title = {Conservative {Agency} via {Attainable} {Utility} {Preservation}},
	isbn = {978-1-4503-7110-0},
	url = {https://dl.acm.org/doi/10.1145/3375627.3375851},
	doi = {10.1145/3375627.3375851},
	urldate = {2024-10-30},
	booktitle = {Proceedings of the {AAAI}/{ACM} {Conference} on {AI}, {Ethics}, and {Society}},
	publisher = {Association for Computing Machinery},
	author = {Turner, Alexander Matt and Hadfield-Menell, Dylan and Tadepalli, Prasad},
	month = feb,
	year = {2020},
	pages = {385--391},
}

@misc{krakovna_penalizing_2019,
	title = {Penalizing {Side Effects} using {Stepwise Relative Reachability}},
	url = {http://arxiv.org/abs/1806.01186},
	language = {en},
	urldate = {2024-10-30},
	publisher = {arXiv},
	author = {Krakovna, Victoria and Orseau, Laurent and Kumar, Ramana and Martic, Miljan and Legg, Shane},
	month = mar,
	year = {2019},
	note = {arXiv:1806.01186 [cs, stat]},
}

@article{kulveit2025gradual,
  title={Gradual {Disempowerment: Systemic Existential Risks} from {Incremental AI Development}},
  author={Kulveit, Jan and Douglas, Raymond and Ammann, Nora and Turan, Deger and Krueger, David and Duvenaud, David},
  journal={arXiv preprint arXiv:2501.16946},
  year={2025}
}

@article{turner2022parametrically,
  title={{Parametrically Retargetable Decision-Makers Tend to Seek Power}},
  author={Turner, Alex and Tadepalli, Prasad},
  journal={Advances in Neural Information Processing Systems},
  volume={35},
  pages={31391--31401},
  year={2022}
}

@inproceedings{guckelsberger2016intrinsically,
  title={{Intrinsically Motivated General Companion NPCs} via {Coupled Empowerment Maximisation}},
  author={Guckelsberger, Christian and Salge, Christoph and Colton, Simon},
  booktitle={2016 IEEE Conference on Computational Intelligence and Games (CIG)},
  pages={1--8},
  year={2016},
  organization={IEEE}
}

@article{van2020robust,
  title={{Robust Multi-Agent Reinforcement Learning with Social Empowerment for Coordination and Communication}},
  author={van der Heiden, Tessa and Salge, Christoph and Gavves, Efstratios and van Hoof, Herke},
  journal={arXiv preprint arXiv:2012.08255},
  year={2020}
}

@inproceedings{woojun_variation_multiagent,
author = {Kim, Woojun and Jung, Whiyoung and Cho, Myungsik and Sung, Youngchul},
title = {A {Variational Approach} to {Mutual Information-Based Coordination} for {Multi-Agent Reinforcement Learning}},
year = {2023},
isbn = {9781450394321},
publisher = {International Foundation for Autonomous Agents and Multiagent Systems},
address = {Richland, SC},
booktitle = {Proceedings of the 2023 International Conference on Autonomous Agents and Multiagent Systems},
pages = {40–48},
numpages = {9},
location = {London, United Kingdom},
series = {AAMAS '23}
}

@article{hammond2025multi,
  title={{Multi-Agent Risks} from {Advanced AI}},
  author={Hammond, Lewis and Chan, Alan and Clifton, Jesse and Hoelscher-Obermaier, Jason and Khan, Akbir and McLean, Euan and Smith, Chandler and Barfuss, Wolfram and Foerster, Jakob and Gaven{\v{c}}iak, Tom{\'a}{\v{s}} and others},
  journal={arXiv preprint arXiv:2502.14143},
  year={2025}
}

@article{song2025estimating,
  title={Estimating the {Empowerment} of {Language Model Agents}},
  author={Song, Jinyeop and Gore, Jeff and Kleiman-Weiner, Max},
  journal={arXiv preprint arXiv:2509.22504},
  year={2025}
}

@inproceedings{mohamed_variational,
author = {Mohamed, Shakir and Rezende, Danilo J.},
title = {Variational {Information Maximisation} for {Intrinsically Motivated Reinforcement Learning}},
year = {2015},
publisher = {MIT Press},
address = {Cambridge, MA, USA},
booktitle = {Proceedings of the 29th International Conference on Neural Information Processing Systems - Volume 2},
pages = {2125–2133},
numpages = {9},
location = {Montreal, Canada},
series = {NIPS'15}
}

@article{carichon2025coming,
  title={The {Coming} {Crisis} of {Multi-Agent Misalignment}: {AI Alignment Must Be a Dynamic and Social Process}},
  author={Carichon, Florian and Khandelwal, Aditi and Fauchard, Marylou and Farnadi, Golnoosh},
  journal={arXiv preprint arXiv:2506.01080},
  year={2025}
}

@inproceedings{chan2023harms,
  title={Harms from {Increasingly Agentic Algorithmic Systems}},
  author={Chan, Alan and Salganik, Rebecca and Markelius, Alva and Pang, Chris and Rajkumar, Nitarshan and Krasheninnikov, Dmitrii and Langosco, Lauro and He, Zhonghao and Duan, Yawen and Carroll, Micah and others},
  booktitle={Proceedings of the 2023 ACM Conference on Fairness, Accountability, and Transparency},
  pages={651--666},
  year={2023}
}

@inproceedings{counterfactual_harm,
author = {Richens, Jonathan G. and Beard, Rory and Thompson, Daniel H.},
title = {Counterfactual {Harm}},
year = {2022},
isbn = {9781713871088},
publisher = {Curran Associates Inc.},
address = {Red Hook, NY, USA},
booktitle = {Proceedings of the 36th International Conference on Neural Information Processing Systems},
articleno = {2634},
numpages = {16},
location = {New Orleans, LA, USA},
series = {NIPS '22}
}

@inproceedings{characterizing_manipulation,
author = {Carroll, Micah and Chan, Alan and Ashton, Henry and Krueger, David},
title = {{Characterizing Manipulation from AI Systems}},
year = {2023},
isbn = {9798400703812},
publisher = {Association for Computing Machinery},
address = {New York, NY, USA},
url = {https://doi.org/10.1145/3617694.3623226},
doi = {10.1145/3617694.3623226},
booktitle = {Proceedings of the 3rd ACM Conference on Equity and Access in Algorithms, Mechanisms, and Optimization},
articleno = {6},
numpages = {13},
location = {Boston, MA, USA},
series = {EAAMO '23}
}

@article{shen2024towards,
  title={{Towards Bidirectional Human-AI Alignment: A Systematic Review for Clarifications, Framework, and Future Directions}},
  author={Shen, Hua and Knearem, Tiffany and Ghosh, Reshmi and Alkiek, Kenan and Krishna, Kundan and Liu, Yachuan and Ma, Ziqiao and Petridis, Savvas and Peng, Yi-Hao and Qiwei, Li and others},
  journal={arXiv preprint arXiv:2406.09264},
  volume={2406},
  pages={1--56},
  year={2024}
}

@misc{salgeEmpowermentIntroduction2013,
    title = {Empowerment -- an {Introduction}},
    url = {http://arxiv.org/abs/1310.1863},
    language = {en},
    urldate = {2024-10-04},
    publisher = {arXiv},
    author = {Salge, Christoph and Glackin, Cornelius and Polani, Daniel},
    month = oct,
    year = {2013},
    note = {arXiv:1310.1863 [nlin]},
}

@article{bakerGoalInferenceInverse2007,
  title = {Goal {{Inference}} as {{Inverse Planning}}},
  author = {Baker, Chris L. and Tenenbaum, J. B. and Saxe, Rebecca R.},
  year = 2007,
  journal = {Proceedings of the Annual Meeting of the Cognitive Science Society},
  volume = {29},
  number = {29},
  urldate = {2026-05-07},
  langid = {english}
}
